\pdfoutput=1
\documentclass[11pt]{article}

\usepackage{ACL2023}

\usepackage{times}
\usepackage{latexsym}

\usepackage[T1]{fontenc}
\usepackage[utf8]{inputenc}

\usepackage{microtype}

\usepackage{inconsolata}

\usepackage{booktabs}       % professional-quality tables
\usepackage{cleveref}
\usepackage{subcaption}
\usepackage{graphicx}

\newcounter{yuanli}
\stepcounter{yuanli}

\title{Trace and Edit Relation Associations in GPT}

\author{%
  Jiahang Li \\ \texttt{jiahang@bu.edu}
  \And 
  Taoyu Chen \\ \texttt{mirack@bu.edu}  
  \And 
  Yuanli Wang \\ \texttt{yuanliw@bu.edu} \\
}

\begin{document}
\maketitle

\begin{abstract}
 This study introduces a novel approach for analyzing and modifying entity relationships in GPT models, diverging from ROME\cite{meng2022locating}'s entity-focused methods. We develop a relation tracing technique to understand the influence of language model computations on relationship judgments. Using the FewRel dataset, we identify key roles of MLP modules and attention mechanisms in processing relationship information. Our method, tested against ROME on a new dataset, shows improved balance in specificity and generalization, underscoring the potential of manipulating early-layer modules for enhanced model understanding and accuracy.

\end{abstract}

\section{Introduction}
Where is the knowledge of entity relations stored in a large language model? Our research suggests that such relations are represented through localized computations in GPT, which can be directly manipulated.

Large language models are capable of predicting factual statements about the world, as noted in a paper by Petroni et al.\cite{petroni-etal-2019-language}. However, while auto-regressive neural networks like GPT are widely used, much remains to be understood about how they store knowledge. Prior studies have mainly focused on how entities convey information, but relations between entities are also significant sources of knowledge. For example, the sentence "The Space Needle is located in the city of Seattle" contains the relation "located in" between the two entities. However, existing methods are inadequate for modifying such relations. Therefore, it is still necessary to investigate how to locate and manipulate relation knowledge within these models. GPT's unidirectional attention and generation capabilities offer further opportunities for novel insights.

To address these challenges, we investigate how entity relations are stored in GPT-like auto-regressive transformer models. Our project milestone involves tracing the effects of hidden state activation within GPT to identify how these models represent relations. Our analysis reveals that MLP modules play a crucial role at the beginning and end, while attention is more important for localizing relations to the final token of the prompt within the facts.

We evaluate our findings by comparing our approach to existing methods in relation extraction (RE) tasks. To evaluate ROME’s impact on more difficult cases, we introduce a dataset of counterfactual assertions
that would not have been observed in pretraining. Our evaluations  confirm
that the fifth layer of MLP can store relation associations that generalize beyond specific surface
forms, while remaining specific to the subject. Compared to original ROME, modified ROME achieves good generalization and specificity simultaneously on the task of editing relation in language models.

\section{Related work}
Numerous approaches have been employed to gain a deeper understanding of the inner workings of language models (LMs). One such method involves training a probing classifier to identify properties within the model's internal representations, though this approach may not capture the network's behavior completely. Another technique focuses on causal effects to extract essential information from the network while avoiding spurious correlations. Some research evaluates LMs' knowledge acquisition by examining their predictive abilities, while other studies attempt to locate and modify knowledge computation within transformers \cite{dai-etal-2022-knowledge}.

The Transformer's multi-head attention mechanism is a crucial component that allows it to attend to various parts of the input sequence simultaneously \cite{vaswani2017attention}. This concept was introduced by dividing the input into multiple heads and applying attention to each head independently, enhancing the model's ability to capture both local and global dependencies.

In recent years, LMs such as BERT, GPT-2, and RoBERTa have achieved notable success in various NLP tasks, including relation extraction (RE). For instance, Cai \cite{yu2019joint} proposes an innovative decomposition strategy for joint entity and relation extraction. Likewise, Yankai Lin \cite{wu2019enriching} introduces a novel method incorporating entity information into pre-trained LMs, which boosts relation classification accuracy.

In the realm of knowledge editing, several investigations have been conducted. Chen et al.\cite{zhu2020modifying} discovered that a straightforward constrained fine-tuning approach, where weights are restricted to remain close to their pre-trained values, is highly effective in modifying the acquired knowledge within a transformer model. Similarly, Nicola De Cao et al.\cite{de2021editing} introduced a "KnowledgeEditor" (KE) hypernetwork, designed to fine-tune a model and integrate new facts provided in the form of textual descriptions. This hypernetwork, an RNN, processes the fact description and the loss gradients to suggest a sophisticated, multilayered modification to the network.

\section{Problem formulation}
Our project focuses on understanding how GPT-like auto-regressive transformer models store entity relations. The project milestone involves analyzing the hidden state activation within GPT to identify how these models represent relations. The findings indicate that MLP modules are crucial at the beginning.

The study evaluates its approach against existing methods in relation extraction tasks and introduces a dataset of counterfactual assertions for more difficult cases. The evaluation confirms that the modified approach can store factual associations that generalize beyond specific surface forms while remaining subject-specific. The modified method achieves good generalization and specificity simultaneously in editing relation tasks in language models.

\section{Methods}
\subsection{Relation Tracing}
Our study is inspired by the process of causal tracing, which we employ to isolate the effect of individual states within a network during statement processing. By tracking the flow of information through the network, we can locate facts within the parameters of a large pretrained auto-regressive transformer.

To pinpoint the hidden states with the most significant impact on predicting the relations of individual facts, we analyze each knowledge tuple t = (s, r, o) containing the subject s, object o, and connecting relation r. We then present a natural language prompt p describing (s, o) to elicit the fact in GPT and examine the model's prediction of r.

The array of states constructs a relation graph that delineates the dependencies among the hidden variables. Our goal is to determine if specific hidden state variables are more crucial than others when recalling a fact. To achieve this, we utilize causal mediation analysis, which quantifies the contribution of intermediate variables in causal graphs (Pearl, 2001).

In order to calculate each state's contribution towards an accurate factual prediction, we observe all of G's internal activations during three runs: a clean run that predicts the fact, a corrupted run with a damaged prediction, and a corrupted-with-restoration run that evaluates the capacity of a single state to restore the prediction.

Our method shares similarities with ROME in terms of executing batches of inferences with two interventions. The first intervention introduces random noise to some batch inputs, while the second intervention transfers clean, non-noised states from an uncorrupted batch member to others. The zeroth element of the batch represents the uncorrupted run, while subsequent elements may be corrupted by providing different input tokens. To guarantee representativeness of corrupted behavior, several (ten) corrupted runs with their unique noise samples are executed within the same batch. By specifying a set of token indices and layers, hidden states can be restored to their values in the uncorrupted run.
\subsection{Editing Relation in GPT}

To enable editing of specific facts within a GPT model, the paper introduces a technique called ROME, or Rank-One Model Editing. This method views an MLP module as a basic key-value store, where the key represents a subject and the value contains information about that subject. Similarly, we apply this method to relation in the language model and change different layers based on the relation tracing results. The MLP can recall the association by retrieving the value associated with the key. This mechinisam employs a rank-one modification of the MLP weights to directly incorporate a new key-value pair.

This model demonstrates a single MLP module within a transformer. The D-dimensional vector functions as the key, representing a relation to be learned about, while the H-dimensional output serves as the value, encoding the relation's acquired attributes. Our method introduces new associations by applying a rank-one alteration to the matrix (d), which maps keys to their corresponding values. According to the results in the relation tracing part, we decided to locate the relaiton in the fifth layer of MLP.

It is important to note that our method adopts a linear perspective of memory within a neural network, as opposed to focusing on individual neurons. This linear approach perceives individual memories as rank-one slices within the parameter space. Experimental evidence supports this viewpoint: when a rank-one update is applied to an MLP module at the computational core identified by causal tracing, it reveals that associations of specific facts can be updated in a manner that is both precise and generalizable.

\section{Dataset}
\subsection{Relation Extraction Migration}
In order to meet our specific requirements for relation extraction, we modified the FewRel dataset\cite{DBLP:journals/corr/abs-1810-10147}, which comprises 100 relations and 70,000 instances sourced from Wikipedia. Each item in the dataset encompasses details about the head, tail, and names. We employed two techniques to reformat the data. First, we harnessed the capabilities of the OpenAI model "text-davinci-003" to generate the necessary templates, predictions, and prompts. Second, we maintained the original context as a prompt while using the same model to create templates, which helped ensure the consistency of predictions and relations with the initial dataset. Lastly, we conducted data cleaning and transformed the dataset to be case-insensitive.

One of the data sample looks like \cref{tab:dataset-item}.

\begin{table*}[ht]
\centering
\caption{Example dataset item from FewRel}
\label{tab:dataset-item}

\begin{tabular}{@{}ll@{}}
\toprule
\textbf{Field} & \textbf{Data Content} \\ \midrule
relation\_triple\_id & 0 \\
subject & tjq \\
relation & territorial entity or entities served by this transport hub (airport, train station, etc.) \\
template &  \"{}, a territorial entity or entities served by this transport hub (airport, train station, etc.)\ \\ 
prediction & territorial entity or entities served by this transport hub (airport, train station, etc.) \\
prompt & merpati flight 106 departed jakarta ( cgk ) on a domestic flight to tanjung pandan ( tjq ) . \\
relaiton\_id & P931 \\
\bottomrule
\end{tabular}
\end{table*}

% {"relation_triple_id": 0, "subject": "tjq", "relation": "territorial entity or entities served by this transport hub (airport, train station, etc.)", "template": " \"{}, a territorial entity or entities served by this transport hub (airport, train station, etc.)\",", "prediction": "territorial entity or entities served by this transport hub (airport, train station, etc.)", "prompt": "merpati flight 106 departed jakarta ( cgk ) on a domestic flight to tanjung pandan ( tjq ) .", "relaiton_id": "P931"}
% 原数据集的每条数据有head，也就是subject，tail，也就是object，names，也就是relation

% 首先生成第一种格式
% 使用openai的模型"text-davinci-003"生成所需要的template，prediction，prompt

% 生成第二种格式
% 使用openai的模型"text-davinci-003"生成所需要的template，
% prompt使用原文
% prediction和relation都是原数据集中的relation

  % fact item 
  % {
  %   "known_id": 0,
  %   "subject": "Vinson Massif",
  %   "attribute": "Antarctica",
  %   "template": "{} is located in the continent",
  %   "prediction": " of Antarctica. It is the largest of the three",
  %   "prompt": "Vinson Massif is located in the continent of",
  %   "relation_id": "P30"
  % },

  % our dataset :
  %   {
  %   "relation_triple_id": 0,
  %   "subject": "A",
  %   "relation": "r",
  %   "template": "{} is B's ",
  %   "prediction": "r",
  %   "prompt": "A is B's"// something prompt to predict their relation,
  %   "relation_id": ""
  % },

\subsection{Dataset for evaluation}
While zero-shot relation extraction (zsRE) metrics in the standard model-editing approach offer a reasonable basis for assessing ROME, they fail to provide in-depth insights that would enable us to differentiate between superficial wording alterations and more profound modifications that reflect a meaningful change in relations. Specifically, we aim to evaluate the effectiveness of significant changes. 

To address this, we assemble a set of more challenging relation-replaced facts (s, r*, o), which initially have lower scores compared to the correct facts (s, r, o). We then introduce the Efficacy Score (ES), which represents the proportion of cases in which P[r*] > P[r] after editing, and the Efficacy Magnitude (EM), which is the average difference between P[r*] and P[r].

First and foremost, it is important to note that there are a total of 64 distinct types of relations. Within this context, we can differentiate between two specific types: 'target true' and 'target new'. The 'target true' relation refers to the original or primary relation present in a given scenario, while 'target new' represents any one of the remaining 63 possible relations.

As our primary concern revolves around understanding and analyzing these relations, the subject matter for our discussion is consistently focused on the term "relation." To examine the connections between different elements or concepts, we utilize a standard prompt, which seeks to explore the relation between two points of interest, labeled as 'a' and 'b' in the given sentence.

In order to gain a comprehensive understanding of the complex relationships between various entities, we consistently pose the following question as our prompt: "What is the relation between 'a' and 'b' in the sentence?" By maintaining a uniform approach to examining relations, we can ensure a systematic and thorough analysis of the wide array of potential connections that may exist between different elements.

\begin{figure*}
    \centering
    \includegraphics[width = 0.9\textwidth]{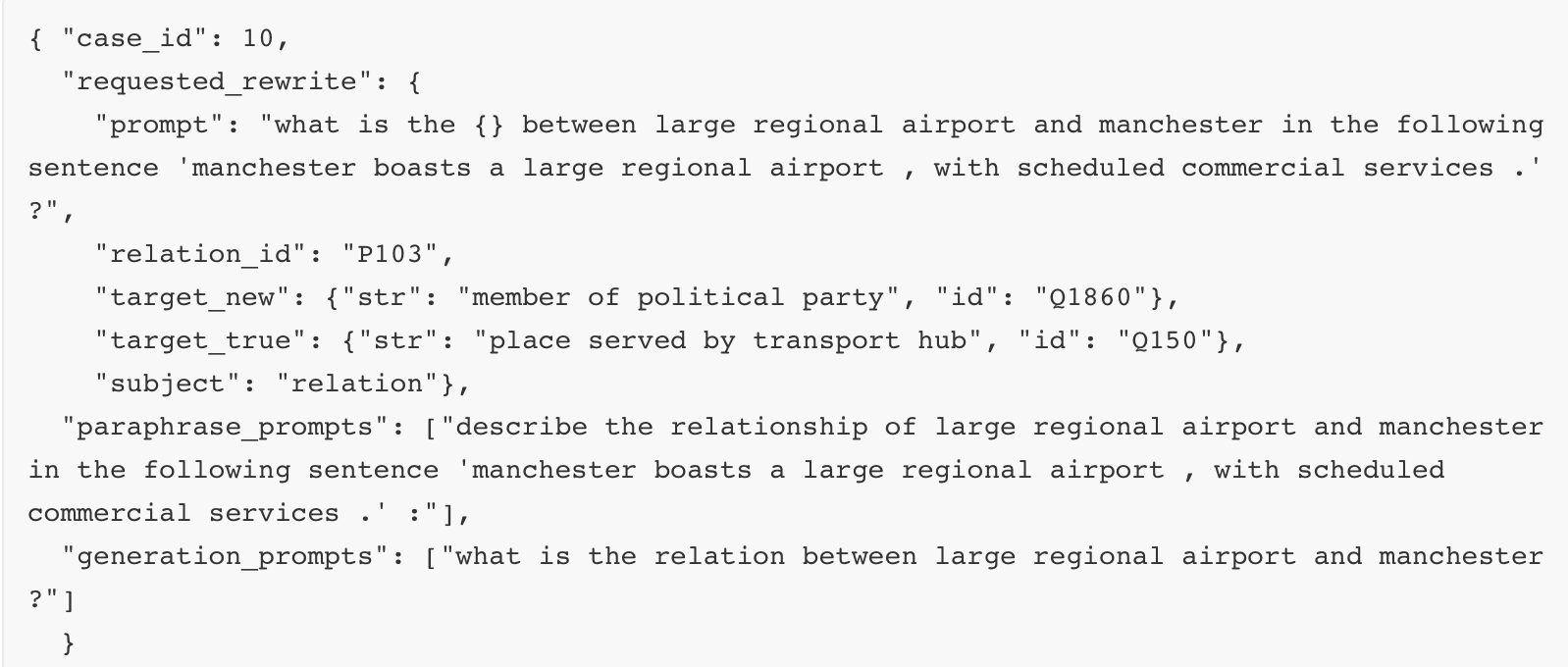}
    \caption{Data item for evaluation on modifying different layer of MLP}
    \label{fig:evaluation}
\end{figure*}

One of the data sample looks like \cref{fig:evaluation}. It aims to identify the correct relation between a large regional airport and Manchester in a given sentence. The focus of the data item is on relations, and it provides a question template, along with the correct relation ("place served by transport hub") and an incorrect relation ("member of political party") for comparison. Additionally, the JSON object includes paraphrased and alternative versions of the question, offering multiple ways to inquire about the relationship between the entities in the sentence.

\section{Results}
\subsection{Confirming the Importance of Decisive States Identified by Relation Tracing}
Our analysis of result heat maps indicates that MLP modules play a crucial role at the beginning and end of the prompt, while attention is more important for localizing relations to the final token of the prompt within the facts. Furthermore, we were able to apply additional operations to the last layer in the MLP module, which is not possible with casual tracing in ROME.

Using relation traces, we propose a specific mechanism for storing relations, which localizes them along three dimensions. This involves placing them in the early sites and last layer of the MLP module and at the processing of the subject's final token.

\begin{figure*}[ht]
\centering

\label{fig:subpictures}

\begin{subfigure}[b]{0.32\linewidth}
    \centering
    \includegraphics[width=\linewidth]{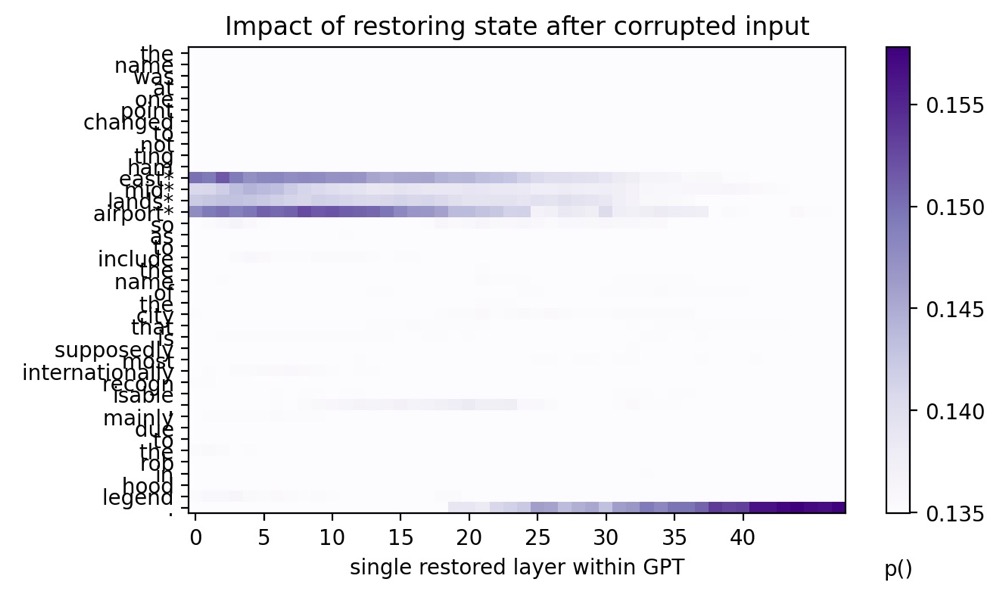}
    \caption{Impact of restoring state after corrupted input}
    \label{fig:sub-a}
\end{subfigure}
\hfill
\begin{subfigure}[b]{0.32\linewidth}
    \centering
    \includegraphics[width=\linewidth]{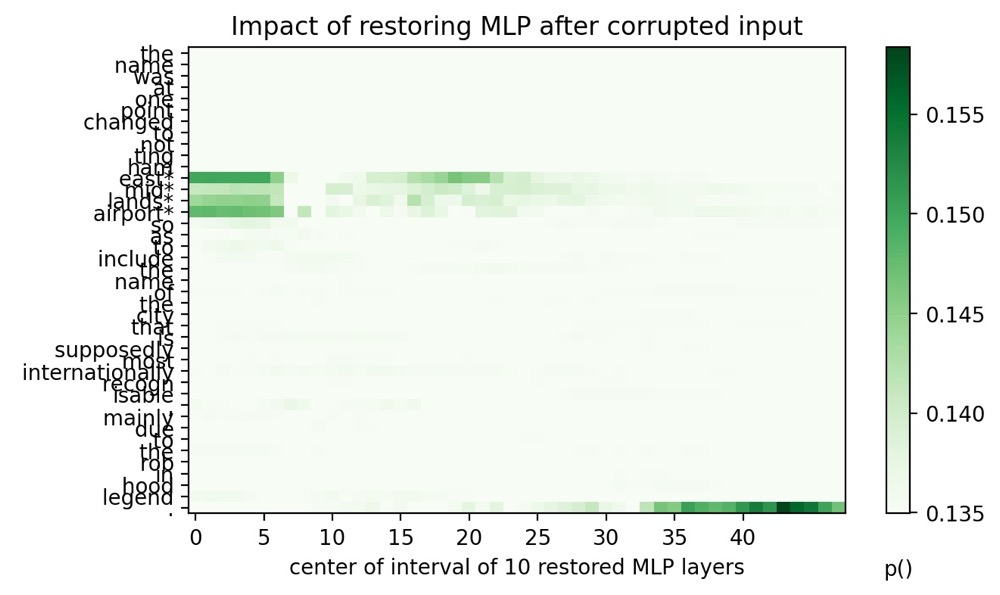}
    \caption{Impact of restoring MLP after corrupted input}
    \label{fig:sub-b}
\end{subfigure}
\hfill
\begin{subfigure}[b]{0.32\linewidth}
    \centering
    \includegraphics[width=\linewidth]{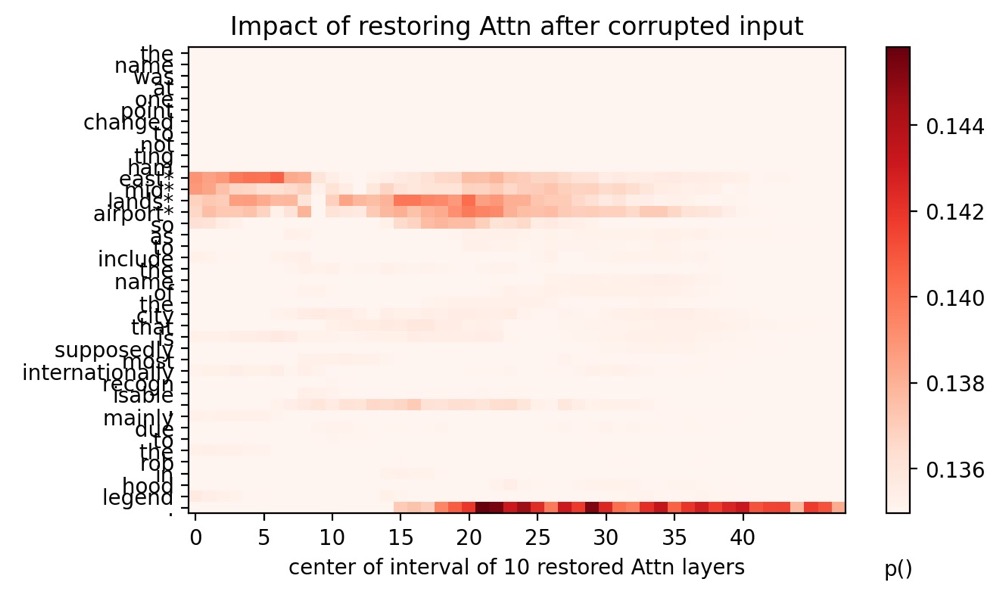}
    \caption{Impact of restoring Attn after corrupted input}
    \label{fig:sub-c}
\end{subfigure}
\caption{The relation impact on output probability}
\end{figure*}

\subsection{Comparing Generation Results}
Initially, we conducted experiments on the original ROME dataset for evaluation purposes. We utilized a subset of the data, consisting of 560 items spanning across 64 distinct relations. The distribution of these relations is depicted in the visualization provided below. By ensuring that our dataset is essentially evenly distributed across the different relations, we aim to mitigate any potential bias that could arise due to imbalances in the representation of individual relations. This balanced distribution allows us to accurately assess the performance of our model and obtain reliable results that can be generalized to a broader range of relations within the ROME dataset. We also mention the most frequent relations in \cref{table:top10_relation_ids}.

\begin{table}[ht]
\centering
\caption{Top 10 Relations for evluation by Frequency}
\begin{tabular}{|c|c|}
\hline
Relation  & Frequency \\
\hline
has part & 18 \\
\hline
instrument & 14 \\
\hline
characters & 14 \\
\hline
mountain range & 13 \\
\hline
country & 13 \\
\hline
screenwriter & 12 \\
\hline
notable work & 12 \\
\hline
occupation & 12 \\
\hline
composer & 12 \\
\hline
located on terrain feature & 12 \\
\hline
\end{tabular}

\label{table:top10_relation_ids}
\end{table}

\begin{figure}
    \centering
    \includegraphics[width = 0.49\textwidth]{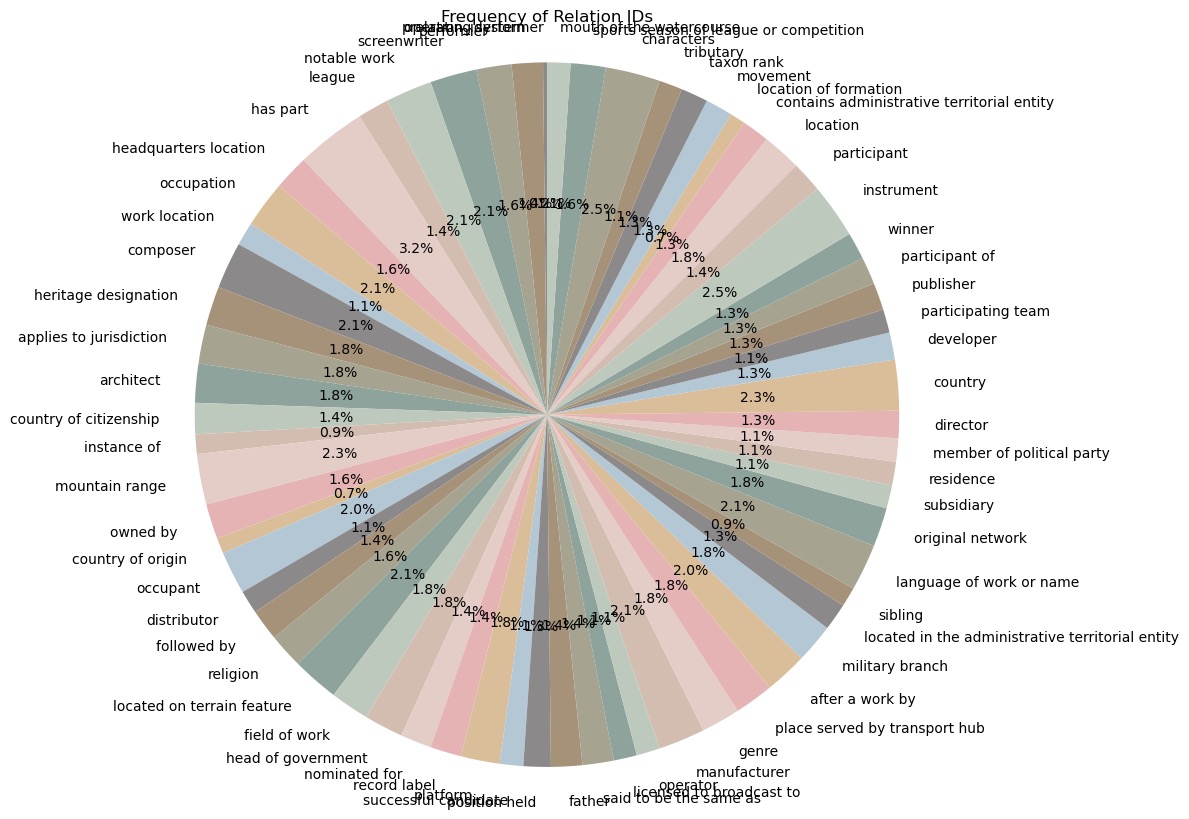}
    \caption{Relation distribution in the dataset for evaluation}
    \label{fig:relation_dis}
\end{figure}

The results presented in \cref{tab:results_summary} showcase the performance and variance of the original MLP, ROME, and our proposed model. Through our analysis, we discovered that directly modifying the key-value pair of layer 5 in the MLP leads to improved performance when altering relation-based knowledge within the GPT model, as compared to the original ROME approach. This improvement is evident in the Paraphrase Success metric, which quantifies the likelihood that the modified model can answer the question according to the altered relation in the language model.

Our proposed model demonstrates a Paraphrase Success probability of 41.07$\%$, outperforming the original ROME model, which has a probability of 40.71$\%$. This result highlights the effectiveness of our model in adapting to changes in relation-based knowledge and underscores its potential for enhancing the capabilities of language models when handling tasks that require an understanding of diverse and dynamic relationships.

\begin{table*}[ht]
\centering
\caption{Metrics if Performance and variance after modifying corresponding layer of MLP in GPT}
\label{tab:results_summary}
\begin{tabular}{lccc}
\toprule
\textbf{Metric}                 & \textbf{Original MLP} & \textbf{Layer 17-ROME} & \textbf{Layer 5-Our Model} \\ \midrule

Paraphrase Diff                 & -0.15 (0.82)        & -0.14 (0.85)      & -0.14 (0.87)      \\
Paraphrase Success              & 39.29 (48.84)       & 40.71 (49.13)     & 41.07 (49.20)     \\
Rewrite Diff                    & -0.05 (0.43)        & 93.00 (15.31)     & 93.67 (13.77)     \\
Rewrite Success                 & 43.04 (49.51)       & 100.00 (0.00)     & 99.82 (4.22)      \\
 \bottomrule
\end{tabular}
\end{table*}

A comparative analysis between modifying the middle layer and the early layer of the MLP reveals that altering the early layer exhibits superior performance in recognizing relation editing and making appropriate adjustments in response to prompted questions. This observation suggests that the early layers of the MLP have a more significant impact on the model's ability to process and adapt to changes in relation-based knowledge. By focusing on the early layers, we can better leverage the model's potential to accurately interpret and respond to questions that involve diverse and dynamic relationships, ultimately enhancing its overall performance in handling complex language tasks.

\begin{figure*}
    \centering
    \includegraphics[width = 0.9\textwidth]{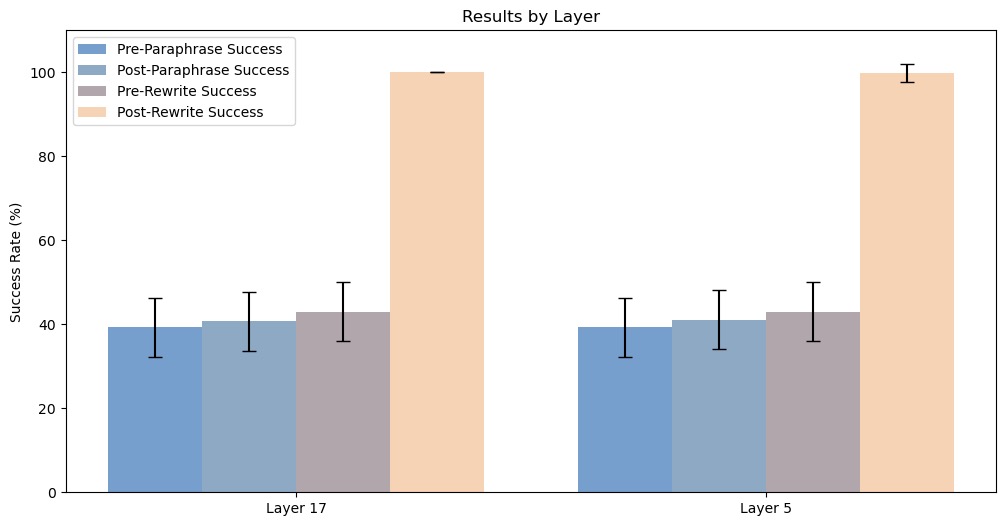}
    \caption{Visualized Performance and variance after modifying corresponding layer of MLP in GPT}
    \label{fig:success}
\end{figure*}

\section{Conclusion}
The presented study offers a significant contribution to the understanding and manipulation of relation stored within large pretrained transformers, such as GPT models. Through the innovative methods of relation tracing and modified ROME, we demonstrate how to isolate the effects of individual states, pinpoint important hidden state variables, and edit relations in the model. 

\section{Future Work}
Modifying the underlying architecture to enable direct editing of relation information could increase the complexity of the model. This might make it harder to train and maintain, as well as potentially slow down inference times. Language is often ambiguous, and the meaning of a relation might change depending on the context. Capturing and storing relation information in a way that accounts for such nuances is challenging for the model. As we found in the relation trace part, there are still different patterns for different relations stored in the language model.

\bibliography{main}
\bibliographystyle{acl_natbib}

% \appendix

% \section{Example Appendix}
% \label{sec:appendix}

% This is a section in the appendix.

\end{document}